%% file: main.tex
\documentclass{article}

\usepackage[preprint]{neurips_2025}

\usepackage[utf8]{inputenc} 
\usepackage[T1]{fontenc}    
\usepackage{hyperref}       
\usepackage{url}            
\usepackage{booktabs}       
\usepackage{amsfonts}       
\usepackage{nicefrac}       
\usepackage{microtype}      
\usepackage{xcolor}  
\usepackage{soul}
\usepackage[most]{tcolorbox}
\usepackage{adjustbox}

\usepackage{graphicx}
\usepackage{subcaption} 
\usepackage{array}
\usepackage{colortbl}
\usepackage[table]{xcolor}
\definecolor{myred}{RGB}{220, 20, 60}

\usepackage{multirow}
\usepackage{multicol}
\usepackage{graphicx}
\usepackage{wrapfig}
\usepackage{algorithm}
\usepackage{algpseudocode}
\usepackage{changepage}

\usepackage{booktabs}
\usepackage{multirow}
\usepackage{makecell}

\usepackage{amssymb, amsmath}
\usepackage{dsfont}
\usepackage{cleveref}



\title{Robo-Cortex: A Self-Evolving Embodied Agent via Dual-Grain Cognitive Memory and Autonomous Knowledge Induction}

\author{%
Nga Teng Chan$^{1,2\diamond}$,
Yi Zhang$^{2\ddagger}$,
Yechi Liu$^{3,2\diamond}$,
Renwen Cui$^{3,2\diamond}$,
Fanhu Zeng$^{3}$,
Zeyuan Ding$^{2}$,\\
\textbf{Xiancong Ren}$^{2}$,
\textbf{Zhang Zhang}$^{3}$,
\textbf{Qifeng Chen}$^{1*}$,
\textbf{Jian Liu}$^{4}$,
\textbf{Yong Dai}$^{2*}$,
\textbf{Xiaozhu Ju}$^{2*}$\\[2mm]
\small
$^{1}$The Hong Kong University of Science and Technology, $^{2}$X-Humanoid, $^{3}$Institute of Automation, CAS\\
\small
$^{4}$Beijing University of Aeronautics and Astronautics\\
\small
$^{*}$Corresponding author,
$^{\ddagger}$Project lead\\
\small
\textbf{Project Website:}
    \href{https://robocortex66.github.io/robo-cortex/}{\textcolor{myred}{https://robocortex66.github.io}}
}

\begin{document}

\maketitle

\begingroup
\renewcommand{\thefootnote}{$\diamond$}
\footnotetext{Work was carried out during the internships of Nga Teng Chan, Yechi Liu, and Renwen Cui at X-Humanoid.}
\endgroup

\vspace{-1.2em}
\begin{center}
  \includegraphics[width=1\textwidth]{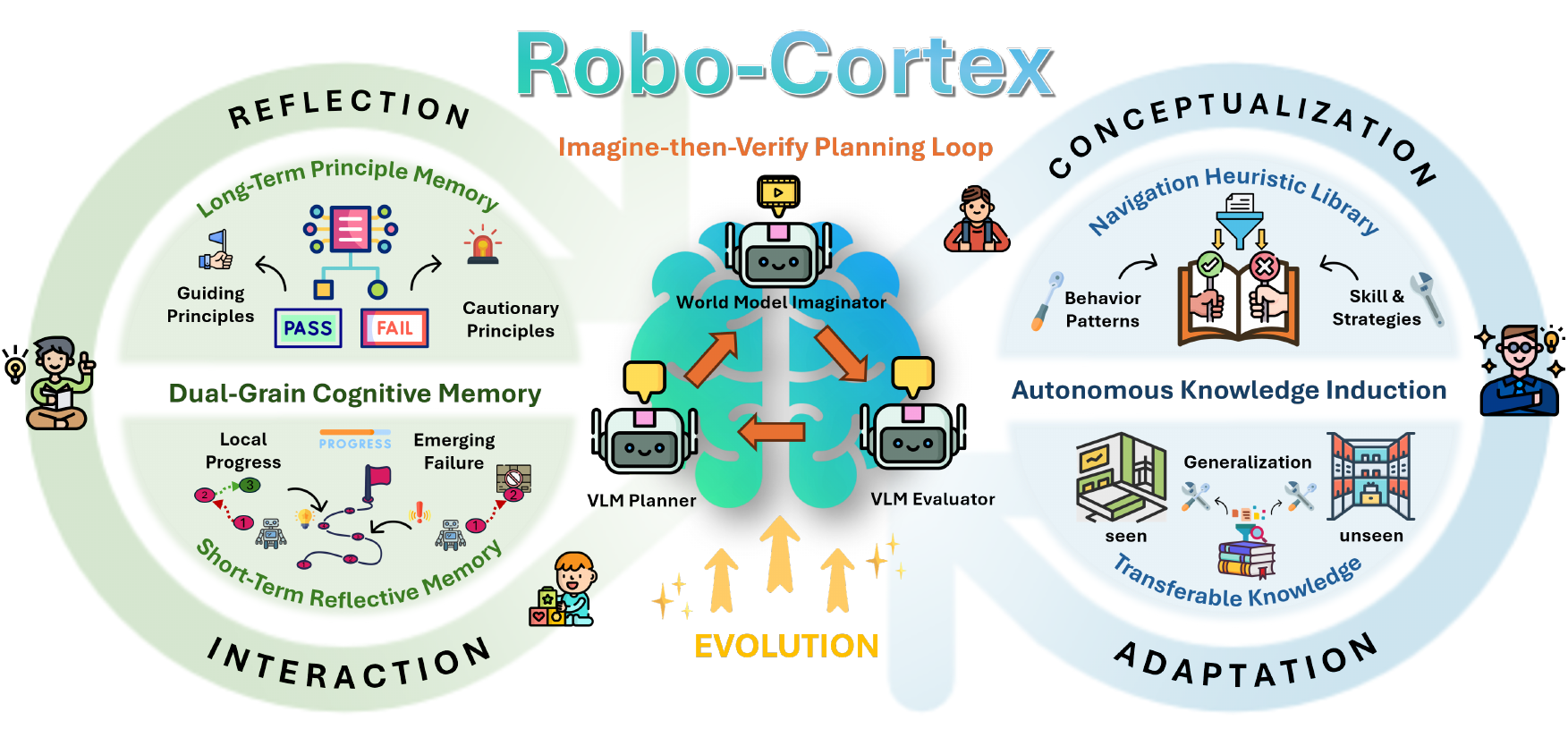}
  \captionof{figure}{\textbf{Overview of Robo-Cortex.} Robo-Cortex is a self-evolving embodied navigation framework with three components: an \emph{Imagine-then-Verify} planning loop for closed-loop decision making, \emph{Dual-Grain Cognitive Memory} for reflection at two temporal scales, and \emph{Autonomous Knowledge Induction} for distilling transferable navigation heuristics from experience. Together, they form an interaction-reflection-conceptualization-adaptation loop for continual strategy evolution.}
  \label{fig:teaser}
\end{center}

\input{0_abstract}    
\input{1_intro}
\input{2_relatedwork}
\input{3_method}

\input{4_expriments}
\input{5_conclusion}

\newpage
\bibliography{main}
\bibliographystyle{plain}

\end{document}

%% file: 0_abstract.tex
\begin{abstract}
\label{sec:abstract}

\begin{tcolorbox}[
    breakable,
    enhanced,
    colback=gray!15,
    colframe=white!0,
    boxrule=0pt,
    arc=4mm,
    left=4mm,
    right=4mm,
    width=\linewidth,
    enlarge left by=0mm,
    enlarge right by=0mm,
    before skip=0pt,
    after skip=0pt
]
The ability to navigate and interact with complex environments is central to real-world embodied agents, yet navigation in unseen environments remains challenging due to “experiential amnesia,” where existing trajectory-driven or reactive policies fail to synthesize generalizable strategies from past interactions. We propose Robo-Cortex, a self-evolving framework that enables robots to autonomously induce navigation heuristics and refine cognitive strategies through a continuous reflection-adaptation loop. By abstracting success patterns and failure pitfalls into natural-language heuristics, Robo-Cortex enables a transition from passive execution to active strategy evolution. Our core innovation is an Autonomous Knowledge Induction (AKI) mechanism that distills multimodal trajectories into a structured Navigation Heuristic Library for knowledge generalization. The architecture further incorporates a Dual-Grain Cognitive Memory system, comprising a Short-term Reflective Memory (SRM) for real-time local progress analysis, and a Long-term Principle Memory (LPM) that abstracts past trajectories into reusable guiding and cautionary principles. To ensure robust decision-making, we introduce a multimodal Imagine-then-Verify loop, where a world model simulates potential outcomes and a VLM-based evaluator validates action plans. Extensive evaluations on IGNav, AR, and AEQA show that Robo-Cortex consistently outperforms strong baselines in both task success and exploration efficiency, with gains of up to +4.16\% SPL over the strongest prior method and up to +15.30\% SPL under heuristic transfer to unseen environments. Preliminary real-world robotic experiments further support the effectiveness of Robo-Cortex in physical settings.
\end{tcolorbox}

\end{abstract}

%% file: 1_intro.tex
\section{Introduction}
\label{sec:introduction}
Autonomous navigation is a foundational capability for real-world embodied agents, which are increasingly needed to carry out exploration and task execution in hazardous, unfamiliar, or dynamic environments \cite{azpurua2023survey, zhang2025embodied}. Achieving this capability, however, requires an agent to continuously integrate visual perception, semantic goals, spatial context, and action consequences under partial observability \cite{gu2022vision}. Despite substantial progress in embodied AI, robust navigation in unseen environments remains challenging \cite{parvaneh2020counterfactual, yu2020take}. Existing methods have improved navigation through stronger planning, richer memory, or more capable reasoning, yet many remain trajectory-driven or reactive: they may imitate successful paths, retrieve episode-specific memories, or select actions from local observations, while still struggling to convert prior interactions into portable decision strategy \cite{anderson2018vision, zhou2024navgpt, zhang2025world, pan2025planning, zhang2025mem2ego, zheng2025esceme}. As a result, experience often improves behavior only weakly and locally, leading to what we term \emph{experiential amnesia}: the failure to transform past interactions into reusable decision knowledge that generalizes across environments.

We envision embodied agents that can evolve their cognition in a human-like manner, improving not merely by accumulating experiences, but by transforming them into better strategies over time. In humans, interaction, reflection, abstraction, and adaptation continuously reshapes human cognition \cite{kolb2014experiential}, turning raw experiences into reusable knowledge for future decision making. Inspired by this process of cognitive evolution, we seek to develop embodied agents that do more than act and remember: they continually transform lived experience into increasingly effective strategies for perception, reasoning, and action, enabling cognition to evolve with every interaction.

In this work, we propose \textbf{Robo-Cortex}, a self-evolving embodied navigation framework that continually transforms accumulated experience into transferable heuristics and refined decision strategies through a reflection-adaptation loop. Robo-Cortex integrates three tightly coupled components. First, it uses an \textbf{Imagine-then-Verify} planning loop that performs short-horizon closed-loop decision making: a world model imagines the likely future outcomes of candidate actions, and a vision-language evaluator verifies which candidate is most promising with respect to the task goal. Second, it introduces a \textbf{Dual-Grain Cognitive Memory} architecture to support adaptation at two complementary time scales. A \textbf{Short-term Reflective Memory} captures local progress, execution context, and emerging failure patterns within an episode, while a \textbf{Long-term Principle Memory} abstracts prior successful and unsuccessful trajectories into reusable guiding and cautionary principles across episodes. Third, Robo-Cortex includes an \textbf{Autonomous Knowledge Induction} (AKI) mechanism that distills recurring multimodal behavior patterns into a structured \emph{Navigation Heuristic Library} for knowledge generalization. In this way, Robo-Cortex goes beyond methods that primarily emphasize planning, memory, or reflection in isolation, and instead unifies multimodal decision-making, reflective memory, and cross-episode heuristic induction within a single embodied framework for continual strategy evolution \cite{shinn2023reflexion, wang2023voyager}.

We evaluate Robo-Cortex on three complementary embodied tasks: image-goal navigation (IGNav), active recognition (AR), and active embodied question answering (AEQA) \cite{zhang2025world}. Across these settings, Robo-Cortex consistently improves both task success and exploration efficiency over strong baselines. We further consider two evaluation regimes: a \textit{static} variant, \textit{Robo-Cortex}, in which memory and heuristic guidance are fixed before testing for fair benchmark comparison, and an \textit{adaptive} variant, \textit{Robo-Cortex++}, in which the agent continues updating its heuristic library online during inference. This separation allows us to distinguish the intrinsic strength of the learned framework from the additional gains enabled by continual self-evolution. Beyond simulation, we also validate Robo-Cortex in real-world robotic deployments, demonstrating its effectiveness in physical environments.

In summary, our contributions are threefold:
\begin{itemize}
    \item We introduce \textbf{Robo-Cortex}, a unified embodied navigation framework that couples short-horizon imagine-then-verify planning with dual-grain cognitive memory to support both online decision-making and experience-aware reasoning.
    \item We propose \textbf{Autonomous Knowledge Induction (AKI)}, which drives the self-evolving capability of Robo-Cortex by inducing reusable heuristics from accumulated experience, updating them online across episodes, and generalizing them to unseen environments.
    \item We demonstrate across three complementary embodied navigation tasks and real-world robot deployments that Robo-Cortex significantly improves success, efficiency, and generalization to unseen environments, validating the benefit of continual heuristic evolution.
\end{itemize}

%% file: 2_relatedwork.tex
\section{Related Work}
\label{sec:relatedwork}

\begin{figure*}
  \includegraphics[width=1\textwidth]{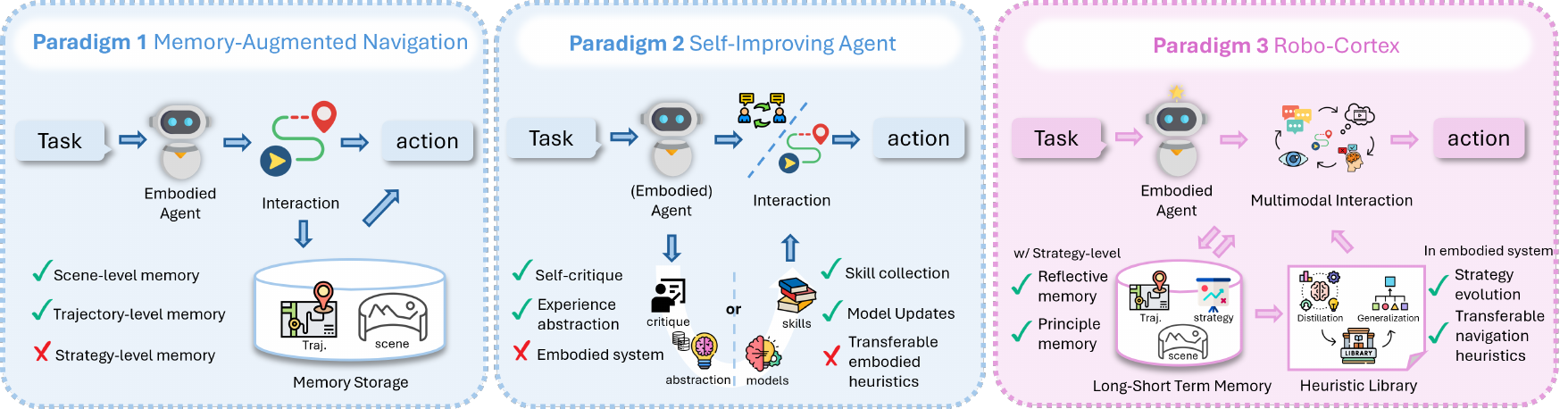}
  \caption{Comparison of prior embodied-agent paradigms and \textbf{Robo-Cortex}. Prior methods typically preserve scene or trajectory context, or improve through critique, skill accumulation, or model updates, but do not explicitly induce transferable embodied heuristics. In contrast, Robo-Cortex forms a self-evolving embodied loop that distills multimodal interaction experience into reusable navigation heuristics for future reflection, planning, and strategy evolution.}
  \label{fig:paradigm}
\end{figure*}

\subsection{Embodied Navigation \& Multimodal Planning}
Embodied navigation studies how an agent perceives, explores, and acts in partially observed environments to reach goals or complete tasks. Early progress was largely driven by training-based policies that map language instructions and egocentric visual observations to actions \cite{anderson2018vision, fried2018speaker, tan2019learning}. Subsequent work enriched this setting with stronger multimodal grounding, trajectory modeling, and spatial reasoning, enabling agents to better integrate visual perception, language, and navigation history for more robust decision making \cite{hao2020towards, chen2021history, chen2022think}. Embodied navigation has also been extended beyond route following to tasks such as active recognition and embodied question answering, where navigation serves not only to move, but also to gather task-relevant evidence through interaction with the environment \cite{das2018embodied, ammirato2017dataset, yokoyama2024hm3d}.

More recently, LLM/VLM-based methods and embodied agents have introduced explicit reasoning, subgoal decomposition, and map-guided or world-model-based planning, substantially improving multimodal decision making within an episode \cite{chen2024mapgpt, zheng2024towards, zhou2024navgpt, zhang2025world, liu2025retrieval}. However, these approaches still primarily optimize action selection under the current episode, whether through better grounding, stronger reasoning, or improved planning modules. By contrast, Robo-Cortex focuses on how embodied experience can be abstracted \emph{across} episodes into reusable navigation heuristics that improve future planning and decision making. In this sense, Robo-Cortex studies embodied navigation not only as a perception and policy problem, but also as a continual strategy-induction problem.

\subsection{Memory for Embodied Navigation}
Memory is central to embodied navigation under partial observability, since the agent must retain information beyond the current observation over long horizons. Prior work therefore equips navigation agents with structured memory mechanisms such as spatial maps, semantic maps, and allocentric memory to preserve explored regions, object cues, and goal-relevant scene structure over time \cite{chaplot2020object, henriques2018mapnet}. Later work extends memory beyond geometric layout to include relational, episodic, and working-memory-style representations that maintain scene structure, object relations, and trajectory context for long-horizon decision making \cite{wu2019bayesian, li2024memonav, yang20253d}. More recent methods further explore memory-efficient long-horizon navigation through annotated semantic maps, retrieved histories, and online map-based memory integrated with world models \cite{zhang2025mapnav, nie2025wmnav, zhang2025mem2ego, zheng2025esceme}.

While these methods substantially improve long-horizon navigation, they primarily retain \emph{scene-level} or \emph{trajectory-level} context. Robo-Cortex instead treats memory as a substrate for reflection and abstraction. Thus, unlike prior scene-centric memory mechanisms, Robo-Cortex uses memory not only to preserve context, but also to induce portable strategy.

\subsection{Self-Improving Agents}
A complementary line of work studies whether agents can improve by reflecting on their own behavior. Recent agentic frameworks show that verbal feedback, self-critique, iterative refinement, and anticipatory introspection can improve future performance without parameter updates \cite{shinn2023reflexion, madaan2023self, wang2024devil}. Subsequent non-embodied systems further demonstrate that prior interactions can be summarized into higher-level guidance that transfers across trials and tasks \cite{park2023generative, zhao2024expel, majumder2023clin, ouyang2025reasoningbank, wu2025evolver}. These works highlight the value of reflection and experience abstraction, but they are typically developed in textual or code-centric settings rather than closed-loop embodied interaction.

Related ideas have also begun to emerge in embodied settings, but with different emphases. Voyager learns a library of executable skills in a simulator-native domain, EvoAgent continually updates the world model for long-horizon embodied tasks, and C-NAV adapts navigation representations to new object categories under continual learning \cite{wang2023voyager, yuan2025evoagent, yu2025cnav}. Other recent navigation works have also explored cross-episode experience accumulation and continual refinement \cite{li2024vision, wang2026lifelong}. Robo-Cortex differs by abstracting multimodal experience into \emph{transferable navigation heuristics} that guide future reflection and planning, generalize to unseen environments, and remain updatable online during inference. This makes Robo-Cortex a framework for continual strategy evolution rather than merely continual refinement.

%% file: 3_method.tex
\section{The Robo-Cortex Framework}
\label{sec:method}

\begin{figure*}
  \includegraphics[width=1\textwidth]{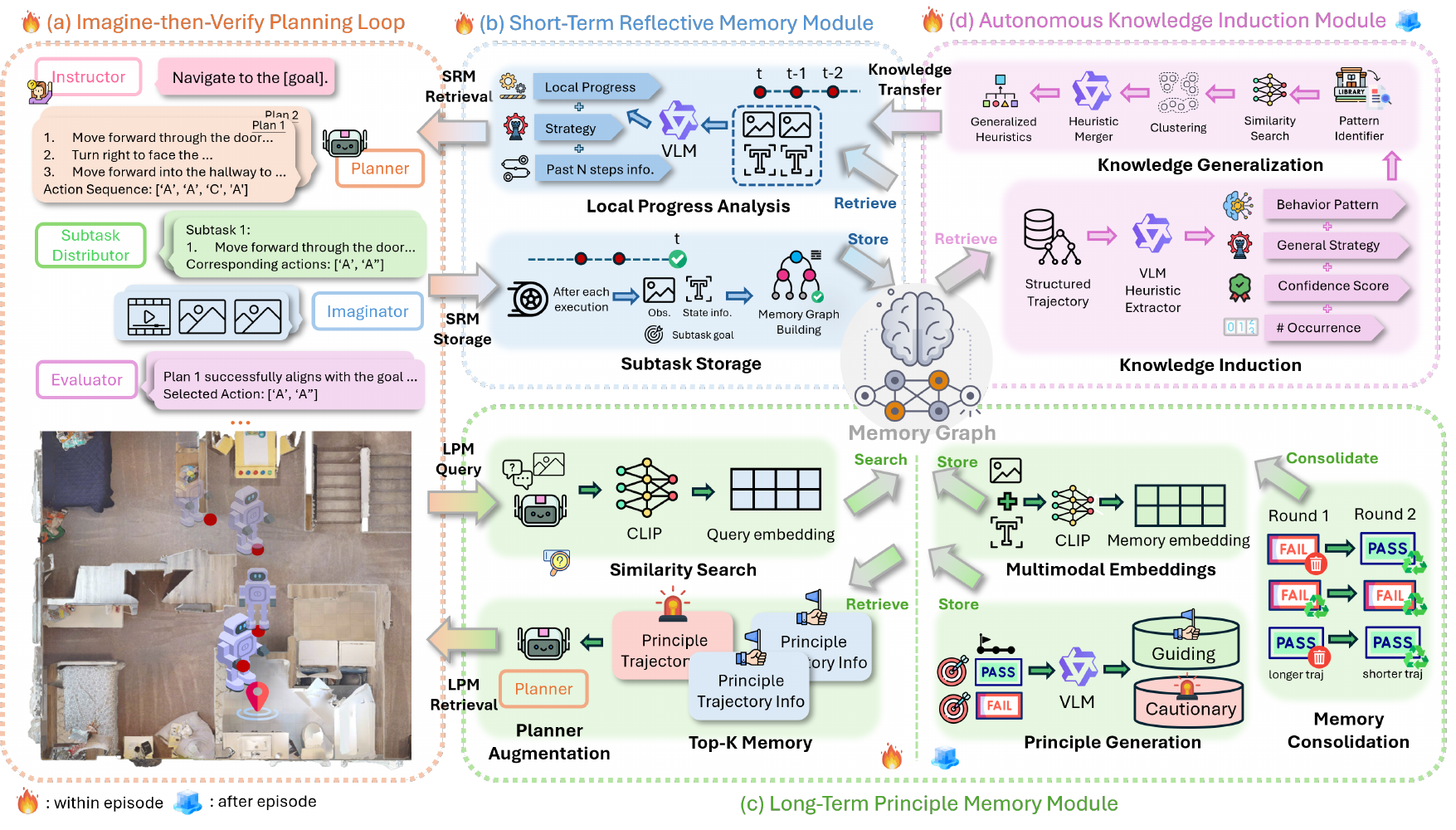}
  \caption{\textbf{Internal Workflow of Robo-Cortex.} Robo-Cortex integrates \textbf{(a) Imagine-then-Verify Planning Loop}, \textbf{(b) Short-Term Reflective Memory}, \textbf{(c) Long-Term Principle Memory} and \textbf{(d) Autonomous Knowledge Induction} through a shared memory graph. During execution, recent subtasks are analyzed for local progress and failure patterns, while related past experiences are retrieved as principle-level guidance. Meanwhile, accumulated trajectories are continually abstracted into reusable navigation heuristics and fed back into future reflection and planning, enabling continual strategy evolution over time.}
  \label{fig:details}
\end{figure*}

\subsection{System Overview}
Robo-Cortex is a self-evolving framework integrating three components: (1) an \textit{Imagine-then-Verify} planning loop for online closed-loop planning, (2) a \textit{Dual-Grain Cognitive Memory} module that combines short-term reflective and long-term principle memory, and (3) an \textit{Autonomous Knowledge Induction} (AKI) module that distills reusable heuristics from accumulated experience.

At each decision step, the agent observes the current state and generates a set of task-conditioned candidate actions or semantic plans. A subtask distributor aligns these candidates with the current semantic goal, after which a world model imagines the future visual outcomes induced by each candidate. A vision-language evaluator then assesses these imagined futures and selects the most promising action or subtask for execution. The selected decision is executed, and the resulting state transition is written to the memory bank. Within each episode, the Short-term Reflective Memory (SRM) module captures local progress and failure patterns to refine subsequent planning in the same trajectory. Across episodes, the Long-term Principle Memory (LPM) module consolidates structured trajectories and their associated principles into persistent knowledge for future planning. On top of this, Autonomous Knowledge Induction (AKI) distills recurring behavioral regularities into reusable heuristics, allowing Robo-Cortex to leverage prior experience as transferable strategic guidance for future planning, and thereby generalize more effectively to unseen environments. Through this design, Robo-Cortex integrates online decision-making, persistent cognitive memory, and continual strategy evolution into a unified embodied navigation system.

\subsection{Imagine-then-Verify Planning Loop}
Since long-range predictions are particularly susceptible to accumulated model error under partial observability, Robo-Cortex avoids committing to extended open-loop action sequences and instead performs short-horizon imagine-select-execute planning. This design supports frequent replanning, allowing the agent to continually revise its decisions as new observations become available.

Let $o_t$ denote the current observation at time step $t$, and let 
$\mathcal{P}_t=\{p_t^{(1)},\dots,p_t^{(N)}\}$ denote the candidate plans proposed by the planner. 
Each candidate plan $p_t^{(i)}$ consists of a low-level action sequence $a_t^{(i)}$ and associated high-level reasoning $r_t^{(i)}$:
\begin{equation}
p_t^{(i)} = \left(a_t^{(i)}, r_t^{(i)}\right).
\end{equation}
Since the granularity of the high-level reasoning may not match that of the low-level action sequence, a subtask distributor aligns the discrete actions with the corresponding semantic rationale(s), producing a sequence of semantically grounded subtask units:
\begin{equation}
\mathcal{S}_t^{(i)} = \mathcal{D}\!\left(a_t^{(i)}, r_t^{(i)}\right),
\end{equation}
where $\mathcal{D}(\cdot)$ denotes the subtask distributor. This action-to-reason alignment ensures that each executable action is associated with an explicit semantic rationale, improving the interpretability of action selection and execution.

For each candidate, the world model predicts a short future rollout,
\begin{equation}
\hat{o}_{t+1:t+h}^{(i)} = \mathcal{W}(o_t, \mathcal{S}_t^{(i)}),
\end{equation}
where $\mathcal{W}(\cdot)$ denotes the world model and $h$ is the imagination horizon. The predicted rollout is then passed to a vision-language evaluator, which scores each candidate according to its expected progress toward the task goal $g$ . The selected plan is:
\begin{equation}
p_t^{*} = \underset{{p_t^{(i)} \in \mathcal{P}_t}}{\arg\max} \ \mathrm{Score}\!\left(\hat{o}_{t+1:t+h}^{(i)}, g\right).
\end{equation}

A key property of this design is frequent replanning. After executing the selected decision, Robo-Cortex updates the state and re-enters the planning loop rather than trusting a long open-loop rollout. This produces a closed-loop decision process that uses imagination locally while remaining responsive to newly observed evidence.

\subsection{Dual-Grain Cognitive Memory}
Robo-Cortex organizes experience as a structured memory graph $\mathcal{G}=(\mathcal{V},\mathcal{E})$, which supports reasoning at two complementary granularities: fine-grained local reflection within an episode and coarse-grained principle retrieval across episodes. The node set is defined as:
\begin{equation}
\mathcal{V}=\mathcal{V}^{\text{root}} \cup \mathcal{V}^{\text{traj}} \cup \mathcal{V}^{\text{sub}},
\end{equation}
where root, trajectory, and subtask nodes represent episodes, decision steps, and executed semantic units, respectively. The edge set includes hierarchical parent-child links and temporal links between consecutive subtasks, enabling both local trajectory analysis and cross-episode memory consolidation.

\subparagraph{\textbf{Short-term reflective memory.}} Short-term reflective memory (SRM) operates over a sliding window of recent subtasks. Let:
\begin{equation}
W_t=\{v_{t-w+1}^{\text{sub}},\dots,v_t^{\text{sub}}\}
\end{equation}
denote the most recent window of subtasks. Once enough subtasks have been accumulated, an SRM analyzer summarizes this local trajectory segment as:
\begin{equation}
s_t^{\text{SRM}} = f_{\text{srm}}(W_t).
\end{equation}
The resulting summary captures local progress, failure patterns, and ongoing subgoal context. This reflective cue is then appended to the planner context for subsequent decisions within the same episode, enabling the agent to preserve short-term behavioral continuity, avoid repeatedly executing ineffective actions, and continue partially completed approaches without replanning the entire strategy at every step.

\subparagraph{\textbf{Long-term principle memory.}} Long-term principle memory (LPM) is constructed after each episode ends. For each subtask, Robo-Cortex traverses the temporal graph forward to collect the downstream trajectory:
\begin{equation}
\tau_i = \{v_i^{\text{sub}}, v_{i+1}^{\text{sub}}, \dots, v_{i+k}^{\text{sub}}\},
\end{equation}
where $k$ is a configured horizon. A principle analyzer converts this trajectory into a guiding principle for successful episodes or a cautionary principle for failed episodes.

At test time, Robo-Cortex retrieves prior subtasks by nearest-neighbor matching in a goal-conditioned state embedding space, retaining only matches above a threshold $\delta_{\text{lpm}}$. 

For each matched subtask, LPM returns not only episodic trajectories but also principle-level abstractions distilled from prior successes and failures, allowing the planner to exploit transferable strategic knowledge rather than merely replaying scene-specific actions. This makes the retrieved memory more robust to environmental variation.

\subparagraph{\textbf{Memory consolidation.}} To maintain a compact yet informative memory over continual self-evolution, Robo-Cortex performs goal-wise consolidation over stored episodes. Successful episodes are prioritized over failed ones, and among successful episodes, shorter trajectories are retained as more efficient behavioral templates. When no successful episode is available, diverse failed episodes are preserved. The rationale is to store the most efficient successful behavior once it is available, while otherwise keeping varied failure trajectories that may reveal complementary failure modes for later principle induction.

\subsection{Autonomous Knowledge Induction (AKI)}
While SRM and LPM enable local reflection and episodic retrieval, they do not by themselves allow the agent to \emph{evolve} its decision strategy over time. AKI addresses this limitation by continuously abstracting the agent's own accumulated interaction history into reusable and generalized behavioral heuristics, and feeding the induced knowledge back into future reflection and planning. In this way, Robo-Cortex moves beyond episodic memory alone and progressively converts past experience into transferable strategic knowledge.

After each episode, the knowledge agent loads the corresponding structured trajectory from the memory graph and applies a vision-language heuristic extractor to identify recurring success and failure patterns. Each extracted heuristic is represented as:
\begin{equation}
h=(\rho,d,a,c,y),
\end{equation}
where $\rho$ is a pattern identifier, $d$ is a problem description, $a$ is a recommended strategy, $c \in [0,1]$ is a confidence score, and $y \in \{\text{success},\text{failure}\}$ indicates whether the heuristic is derived from successful or failed behavior.

As new episodes accumulate, the extracted heuristics are merged into a shared knowledge base. Heuristics are first grouped by pattern identifier, and optionally further refined by semantic similarity over their textual descriptions and recommended strategies. For each resulting cluster $\mathcal{C}$, a merger model produces a generalized heuristic:
\begin{equation}
\tilde{h}_{\mathcal{C}}=(\rho,\tilde{d},\tilde{a},\tilde{c}),
\end{equation}
where $\tilde{d}$ and $\tilde{a}$ denote the merged problem description and strategy, and $\tilde{c}$ is the aggregated confidence. Through this process, Robo-Cortex incrementally refines its heuristic library over time, moving from isolated episodic traces to compact behavioral knowledge supported by repeated evidence across episodes.

Unlike LPM, which retrieves principle memories from specific matched experiences, AKI performs cross-episode abstraction to induce shared heuristics that are not tied to any single trajectory. As a result, the induced knowledge is more transferable than raw action traces or scene-specific memories, and can provide strategic guidance that generalizes across varied situations, including unseen environments.

Finally, high-confidence recurrent heuristics are fed back into the online system as structured guidance for short-term reflective analysis and planning. These heuristics provide priors on likely failure modes and promising behaviors, helping the agent interpret recent interaction history more effectively and produce more informative reflective summaries for subsequent decisions. This closes the self-evolving loop of Robo-Cortex: experience is accumulated through interaction, abstracted into reusable knowledge, and then reused to improve future behavior.

%% file: 4_expriments.tex
\section{Experiments}
\label{sec:experiments}
\begin{table}[t]
  \caption{Comparison with state-of-the-art methods on IGNav, AR, and AEQA. Robo-Cortex is evaluated in both static and adaptive settings for each task.}
  \label{tab:main_results}
  \centering
  \small
  \setlength{\tabcolsep}{4.5pt}
  \begin{adjustbox}{max width=\textwidth}
  \begin{tabular}{lcccccccc}
    \toprule
    \multirow{2}{*}{\textbf{Method}} 
      & \multicolumn{3}{c}{\textbf{IGNav}} 
      & \multicolumn{2}{c}{\textbf{AR}} 
      & \multicolumn{3}{c}{\textbf{AEQA}} \\
    \cmidrule(lr){2-4}
    \cmidrule(lr){5-6}
    \cmidrule(lr){7-9}
      & SR$\uparrow$ 
      & SPL$\uparrow$ 
      & Mean Traj.$\downarrow$
      & SR$\uparrow$ 
      & Mean Traj.$\downarrow$
      & Score$\uparrow$
      & Mean Traj.$\downarrow$
      & SPL$\uparrow$ \\
    \midrule
    MapGPT \cite{chen2024mapgpt}       & 17.36 & 15.11 & 56.48 & 10.14 & 9.19 & 25.50 & 18.58 & 22.40 \\
    WMNav \cite{nie2025wmnav}          & 26.39 & 23.36 & 50.60 & 12.96 & 8.89 & 19.50 & 19.98 & \underline{23.60} \\
    World-In-World \cite{zhang2025world} & 38.57 & 27.50 & \textbf{47.10} & 20.68 & \underline{7.09} & 27.19 & 20.53 & 20.56 \\
    \textbf{Robo-Cortex}               & \underline{41.26} & \underline{31.66} & 49.90 
                 & \underline{22.39} & \textbf{6.97} 
                 & \underline{29.78} & \underline{17.34} & 21.40 \\
    \textbf{Robo-Cortex++}             & \textbf{45.07} & \textbf{35.06} & \underline{47.50} 
                 & \textbf{23.88} & 7.57 
                 & \textbf{30.59} & \textbf{16.61} & \textbf{25.57} \\
    \bottomrule
  \end{tabular}
  \end{adjustbox}
\end{table}

\subsection{Experimental Setup}
\subparagraph{\textbf{Datasets and evaluation metrics.}}
We evaluate Robo-Cortex on three embodied tasks following~\cite{zhang2025world}: image-goal navigation (IGNav), active recognition (AR), and active embodied question answering (AEQA). In particular, IGNav and AEQA are evaluated on Habitat-based episodes built on the HM3D validation split \cite{ramakrishnan2021habitat}, while AR is evaluated on Habitat-based episodes built on MP3D \cite{chang2017matterport3d}. Together, these benchmarks span complementary forms of embodied navigation and provide a broad test of generalization across task settings. 

For \textbf{IGNav}, the agent navigates to a target location specified by a goal image. We report \textit{Success Rate} (SR), \textit{Success weighted by Path Length} (SPL), and mean trajectory length.
For \textbf{AR}, the agent explores the scene to recognize the target object category. We report SR and mean trajectory length.
For \textbf{AEQA}, the agent explores the environment to answer a scene-grounded natural-language question. We report the protocol-defined \textit{Answer Score}, SPL, and mean trajectory length.

\subparagraph{\textbf{Baselines.}}
We compare Robo-Cortex with three recent strong baselines: MapGPT~\cite{chen2024mapgpt}, WMNav~\cite{nie2025wmnav}, and World-In-World~\cite{zhang2025world}. They represent competitive embodied agents with different emphases on map-guided planning, world-model-based navigation, and closed-loop embodied decision making.

\subparagraph{\textbf{Implementation details.}} All simulation experiments are conducted in Habitat-sim \cite{savva2019habitat}. We use \texttt{Qwen2.5-VL-72B-Instruct-AWQ} \cite{bai2025qwen2} as the shared vision-language backbone for planning, memory analysis, and AKI, and \texttt{Wan2.1-I2V-A14B-480P-Diffusers} \cite{wan2025wan} as the backbone for world-model imagination. For efficiency, planning and memory-related modules are deployed as separate vLLM services, while sharing the same backbone weights. Moreover, the navigation action space is discretized into a forward motion of 0.20\,m and left/right rotations of $22.5^\circ$. Additional implementation details are provided in the supplementary material.

\begin{table*}[t]
\caption{Memory accumulation study on the seen split. We compare the Basic Pipeline and Dual-Grain Memory variants across multiple rounds to evaluate how progressive experience accumulation improves performance.}
\label{tab:ablation1}
\centering
\footnotesize
\setlength{\tabcolsep}{3.5pt}
\renewcommand{\arraystretch}{1.08}

\begin{tabular}{@{}l c ccc cc ccc@{}}
\toprule
\multirow{2}{*}{\textbf{Method}} 
& \multirow{2}{*}{\textbf{Round}}
& \multicolumn{3}{c}{\textbf{IGNav}} 
& \multicolumn{2}{c}{\textbf{AR}} 
& \multicolumn{3}{c}{\textbf{AEQA}} \\
\cmidrule(lr){3-5}
\cmidrule(lr){6-7}
\cmidrule(lr){8-10}
& 
& SR$\uparrow$
& SPL$\uparrow$
& Mean Traj.$\downarrow$
& SR$\uparrow$
& Mean Traj.$\downarrow$
& Score$\uparrow$
& Mean Traj.$\downarrow$
& SPL$\uparrow$ \\
\midrule

Basic Pipeline 
& \multirow{2}{*}{1} 
& 36.11 & 28.16 & 50.90 
& 20.93 & 6.61 
& 25.55 & 19.22 & 18.55 \\

+ SRM          
&                    
& 40.18 & 30.25 & 50.05 
& 20.93 & 6.22 
& 30.83 & 18.80 & 23.77 \\

\midrule

+ LPM          
& \multirow{2}{*}{2} 
& 37.50 & 28.74 & 50.90 
& 21.71 & 6.55 
& 25.00 & 19.46 & 18.75 \\

+ LPM \& SRM   
&                    
& 42.65 & 32.54 & 48.90 
& 20.16 & 6.26 
& \textbf{33.70} & 17.50 & 21.12 \\
\midrule
+ LPM          
& \multirow{2}{*}{3} 
& 43.06 & \textbf{34.94} & \textbf{47.30} 
& 20.97 & \textbf{6.03} 
& 28.80 & 18.71 & 21.66 \\

+ LPM \& SRM   
&                     
& \textbf{44.29} & 34.74 & 47.90 
& \textbf{24.03} & 6.18 
& 31.71 & \textbf{17.15} & \textbf{27.51} \\

\bottomrule
\end{tabular}
\end{table*}

\begin{table*}[t]
\caption{Heuristic transfer study on the unseen split. We compare manual guidance, online updating, transferred heuristics, and transferred heuristics with continued online updating to evaluate how induced heuristics generalize and adapt in novel environments. Here, heur. denotes heuristics}
\label{tab:ablation2}
\centering
\footnotesize
\setlength{\tabcolsep}{3.5pt}
\renewcommand{\arraystretch}{1.08}

\begin{tabular}{@{}l ccc cc ccc@{}}
\toprule
\multirow{2}{*}{\textbf{Method}} 
& \multicolumn{3}{c}{\textbf{IGNav}} 
& \multicolumn{2}{c}{\textbf{AR}} 
& \multicolumn{3}{c}{\textbf{AEQA}} \\
\cmidrule(lr){2-4}
\cmidrule(lr){5-6}
\cmidrule(lr){7-9}
& SR$\uparrow$
& SPL$\uparrow$
& Mean Traj.$\downarrow$
& SR$\uparrow$
& Mean Traj.$\downarrow$
& Score$\uparrow$
& Mean Traj.$\downarrow$
& SPL$\uparrow$ \\
\midrule

Basic Pipeline 
& 34.72 & 24.03 & 48.50 
& 19.42 & 7.91 
& 32.42 & 19.06 & 26.80 \\

w/ manual heur.
& 38.89 & 30.96 & 50.10 
& 18.71 & 7.82 
& 31.59 & 18.29 & 27.04 \\

w/ transferred heur.
& \textbf{48.61} & \textbf{39.33} & \textbf{46.20} 
& \textbf{23.02} & 7.78 
& 32.88 & 19.47 & 24.31 \\

w/ seen heur. + update 
& 41.67 & 31.69 & 49.50 
& 19.42 & \textbf{7.58} 
& 33.06 & 18.51 & 26.69 \\

w/ manual heur. + update 
& 36.62 & 28.06 & 50.70 
& 22.30 & 7.71 
& \textbf{35.33} & 17.97 & \textbf{28.47} \\

w/ transferred heur. + update 
& 41.67 & 34.84 & 48.30 
& \textbf{23.02} & 7.80 
& 32.22 & \textbf{17.95} & 25.50 \\

\bottomrule
\end{tabular}
\end{table*}

\subsection{Comparison with SOTA Methods}
We compare Robo-Cortex with prior state-of-the-art methods \cite{chen2024mapgpt, nie2025wmnav, zhang2025world}. Robo-Cortex denotes the full \textit{static} system with fixed heuristic guidance during evaluation, while Robo-Cortex++ denotes the \textit{adaptive} variant that updates heuristics online during inference. We report this variant additionally to show the gains enabled by continual self-evolution beyond a fixed policy.

As shown in Table~\ref{tab:main_results}, Robo-Cortex consistently outperforms prior methods across all three embodied tasks. Under the static setting, it already achieves the strongest overall performance, reaching 41.26\% SR and 31.66\% SPL on IGNav, 22.39\% SR on AR, and 29.78 Answer Score on AEQA. Compared with the strongest baseline, Robo-Cortex improves IGNav by +2.69\% SR and +4.16\% SPL, while also achieving the best AR success rate and the shortest trajectories on both AR and AEQA. These results indicate that Robo-Cortex improves not only task success, but also exploration quality and decision efficiency.

The adaptive variant further highlights the self-evolving nature of the framework. With online heuristic updating during inference, Robo-Cortex++ improves the static model on all three tasks, reaching 45.07\% SR on IGNav, 23.88\% SR on AR, and 30.59 Answer Score with 25.57\% SPL on AEQA. This shows that Robo-Cortex is not only a stronger static embodied agent, but also a framework whose strategy can continue to improve through online refinement.

\begin{figure*}[t]
    \centering
    \begin{subfigure}[t]{0.32\textwidth}
        \centering
        \includegraphics[width=\linewidth]{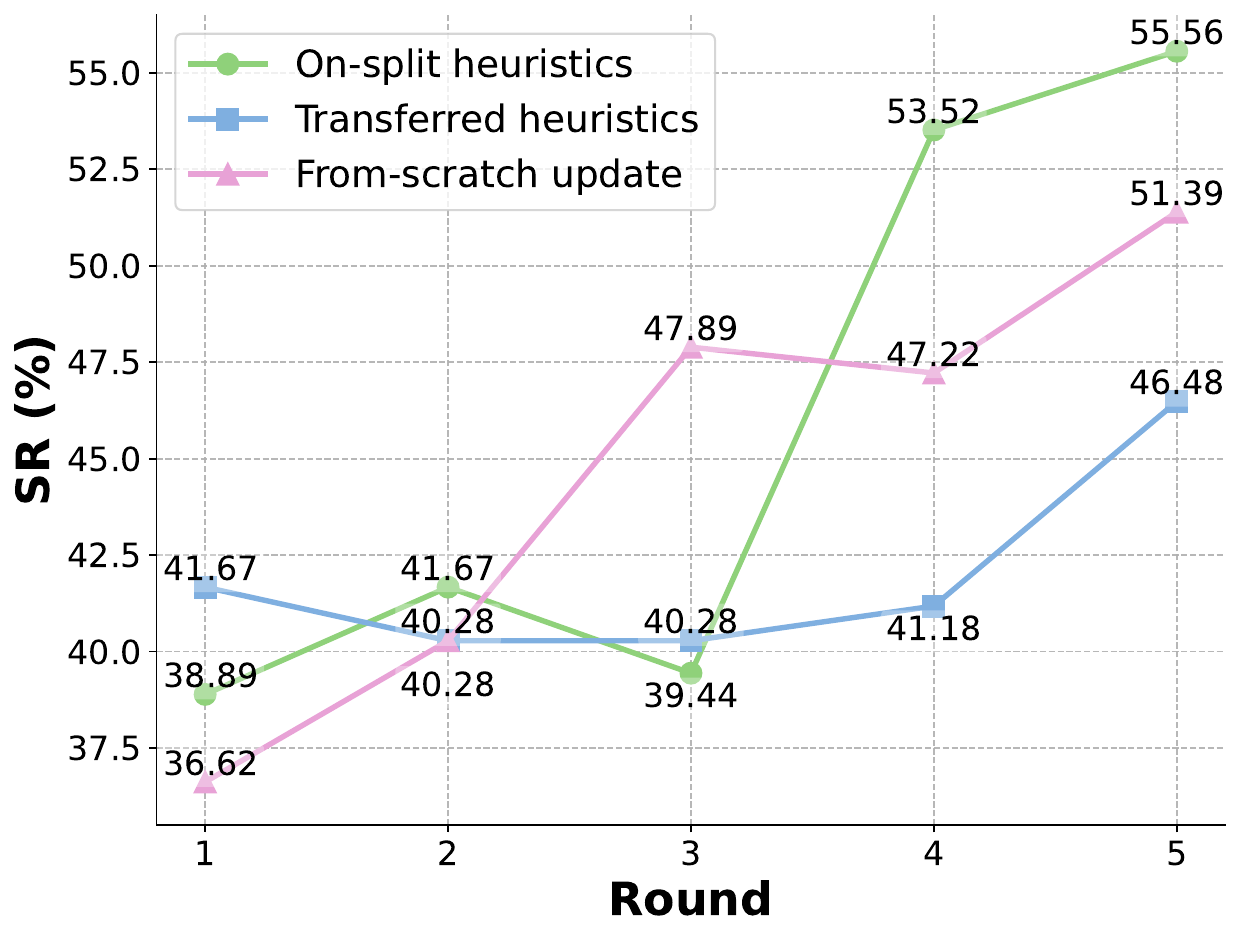}
        \caption{Success Rate (SR)}
        \label{fig:ignav_sr}
    \end{subfigure}\hfill
    \begin{subfigure}[t]{0.32\textwidth}
        \centering
        \includegraphics[width=\linewidth]{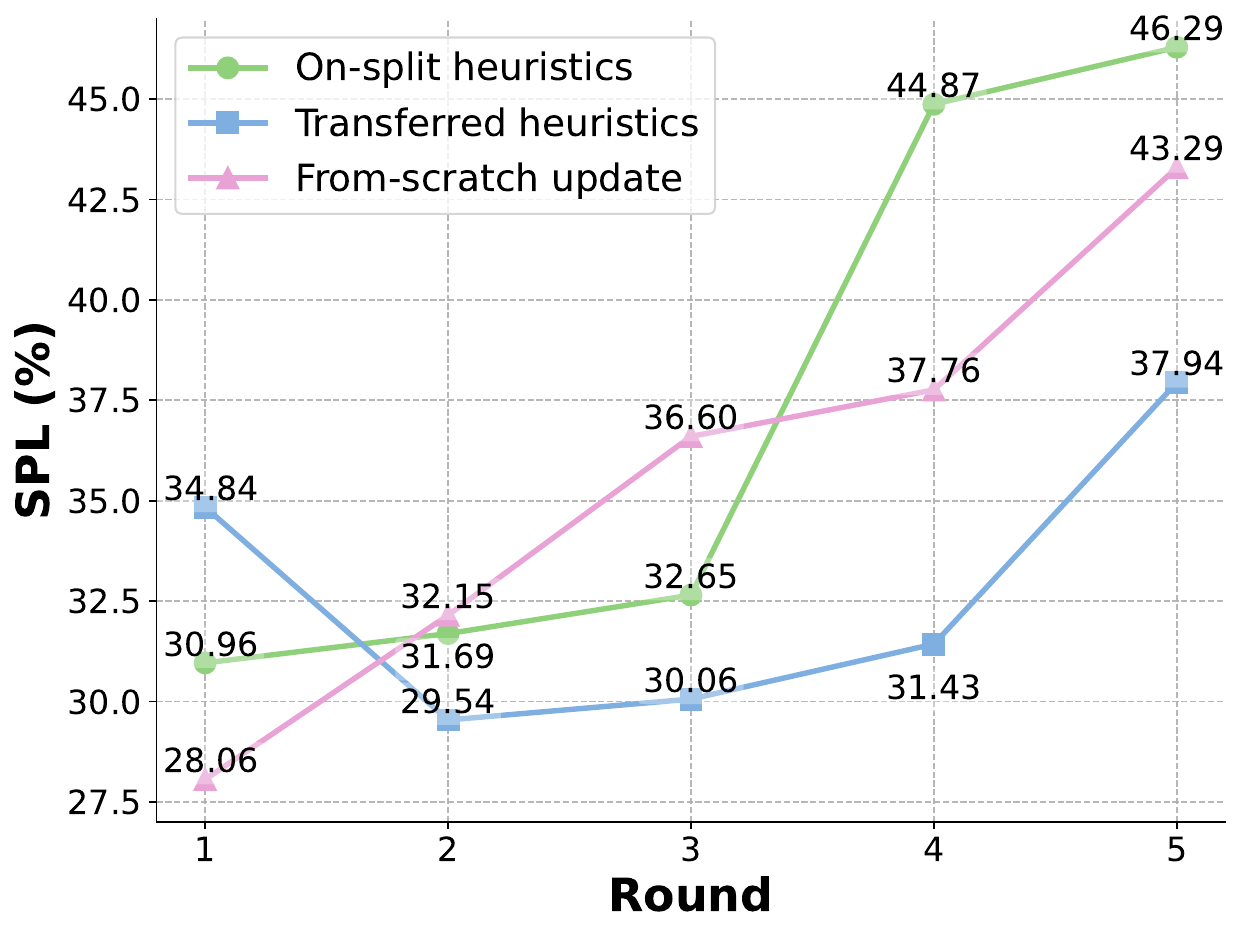}
        \caption{Success weighted by Path Length (SPL)}
        \label{fig:ignav_spl}
    \end{subfigure}\hfill
    \begin{subfigure}[t]{0.32\textwidth}
        \centering
        \includegraphics[width=\linewidth]{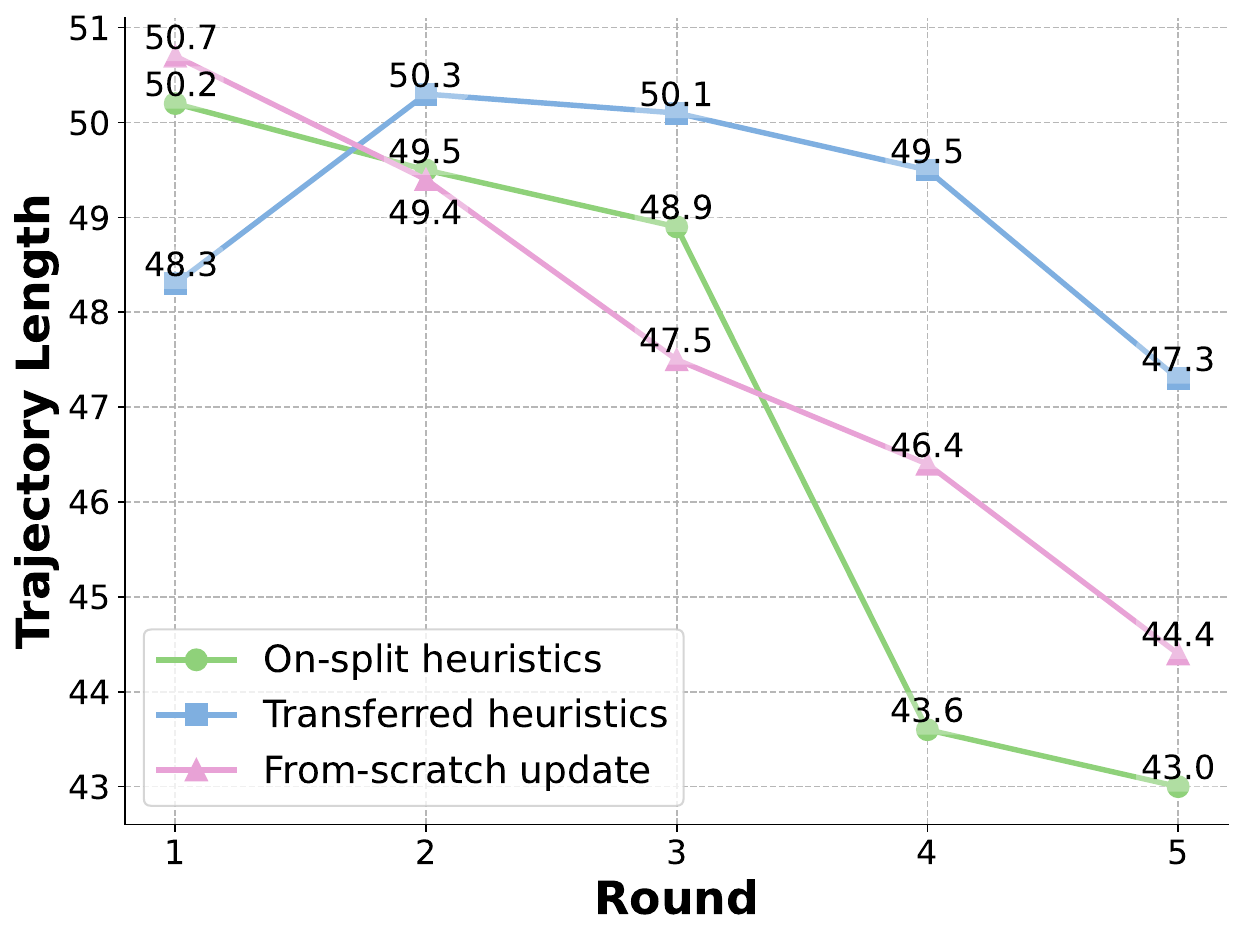}
        \caption{Mean trajectory length}
        \label{fig:ignav_traj}
    \end{subfigure}
    \caption{\textbf{Round-wise update dynamics on IGNav.} We visualize how updating over multiple rounds affects IGNav under three initialization settings: on-split heuristics, transferred heuristics, and from-scratch update. Transferred heuristics provide the strongest initial prior, while continued updating on the target split yields larger long-term gains in both success and efficiency.}
    \label{fig:ignav_update_dynamics}
\end{figure*}

\subsection{Ablation Study}
We separate the evaluation set into two disjoint halves, \textit{i.e.}, seen and unseen, to comprehensivly evaluate Robo-Cortex. On the seen split, we analyze how SRM and LPM improve performance as experience accumulates over multiple rounds. On the unseen split, we evaluate whether heuristics extracted from prior experience transfer to novel episodes, and whether further online updating provides additional gains. As a reference, we define a \textbf{Basic Pipeline} that contains only the core Imagine-then-Verify loop, including planning, world-model imagination, evaluation, and execution, while excluding SRM, LPM, and AKI.

\subparagraph{\textbf{Seen split.}}
We first analyze the contribution of Dual-Grain Cognitive Memory under repeated evaluation on the same split. Since SRM operates within an episode whereas LPM depends on accumulated past experience, we report the Basic Pipeline and \textbf{+SRM} in round 1, and evaluate LPM-based variants after repeated rounds. As shown in Table~\ref{tab:ablation1}, \textbf{SRM} already provides clear first-round gains, improving IGNav from 36.11\% to 40.18\% SR and AEQA from 25.55 to 30.83 Answer Score, while also shortening the AR trajectory. In contrast, the benefit of \textbf{LPM} becomes more pronounced as experience accumulates: its gains are modest at round 2 but substantially stronger by round 3, where it reaches 43.06\% SR on IGNav and improves AEQA to 28.80 Answer Score with shorter trajectories. The full \textbf{+LPM \& SRM} model yields the strongest and most consistent overall results, achieving 44.29\% SR on IGNav, 24.03\% SR on AR, and 31.71 Answer Score on AEQA at round 3. These results support the complementary roles of the two memory components: SRM improves within-episode reflection and failure recovery, while LPM provides increasingly useful cross-episode strategic guidance as experience accumulates.

\subparagraph{\textbf{Unseen split.}}
We next isolate heuristic transfer on novel episodes. In this setting, no LPM is used, so the comparison focuses on heuristic guidance and online updating. As shown in Table~\ref{tab:ablation2}, manually written heuristics provide a useful prior, improving IGNav from 34.72\% to 38.89\% SR, but their benefits are less consistent on AR and AEQA. In contrast, heuristics transferred from the seen split generalize strongly to unseen environments: without any further updating, \textbf{w/ transferred heuristics} achieves the best IGNav performance (48.61\% SR) and the highest AR success rate (23.02\%), showing that AKI captures portable success and failure patterns rather than split-specific trajectories. Online updating on the unseen split yields a more nuanced effect. While \textbf{w/ transferred heuristics + update} still outperforms the Basic Pipeline on IGNav and AR, it does not surpass the fixed transferred-heuristic setting, suggesting that strong transferred heuristics already provide a high-quality prior and that aggressive online rewriting on limited unseen episodes may partially overwrite useful guidance.  Overall, the unseen-split results validate the central role of AKI: heuristics induced from prior experience are transferable to novel environments, and their main value lies in supplying compact strategic priors that complement, rather than replace, online adaptation.

\subparagraph{\textbf{Round-wise update dynamics on IGNav.}}
To further analyze how online updating changes performance over time, we visualize the round-wise dynamics on IGNav in Fig.~\ref{fig:ignav_update_dynamics}. This figure should be viewed as a case study of update behavior, while Tables~\ref{tab:ablation1} and~\ref{tab:ablation2} remain the main multi-task ablation results. The transferred setting provides the strongest initial prior, achieving 41.67\% SR and 34.84\% SPL in round 1, which confirms that heuristics induced from prior experience are immediately useful on unseen episodes. However, its gains plateau quickly, then reaching 46.48\% SR by round 5, with trajectory length 47.3. In contrast, from-scratch updating improves more steadily, increasing from 36.62\% SR in round 1 to 51.39\% in round 5, while shortening the trajectory from 50.7 to 44.4. The strongest long-run performance is obtained by on-split heuristics, which rise from 38.89\% SR and 30.96\% SPL in round 1 to \textbf{55.56\%} SR and \textbf{46.29\%} SPL in round 5, together with a trajectory reduction from 50.2 to 43.0. Notably, this setting shows a sharp improvement after round 3, suggesting that once sufficient target-split experience has accumulated, the induced heuristics become substantially more effective. Overall, these curves reveal a clear trade-off between \emph{fast transfer} and \emph{long-term adaptation}: transferred heuristics offer the best early prior, whereas repeated updating on the target split ultimately yields the strongest navigation performance.

\begin{figure*}
  \includegraphics[width=1\textwidth]{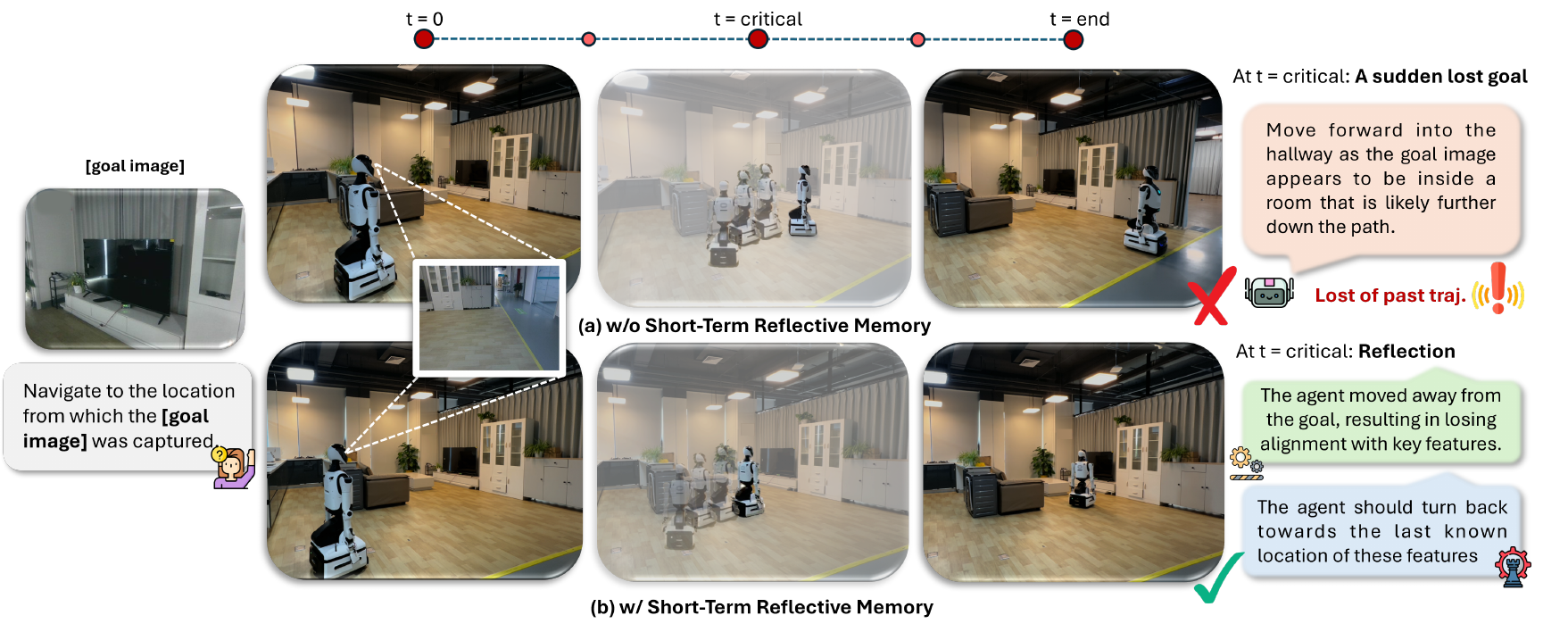}
  \caption{\textbf{Real-world benefit of short-term reflection.} In an image-goal navigation task, the robot without SRM drifts away from the target after losing goal-relevant cues at a critical step. With SRM, Robo-Cortex detects the misalignment, reflects on the failure, and recovers by returning toward the last known goal-consistent region, leading to successful completion.}
  \label{fig:real_world}
\end{figure*}

\subsection{Real-World Experiment}
We conduct a preliminary real-world study on image-goal navigation (IGNav) in two indoor scenes (i.e. living room and supermarket). The Basic Pipeline succeeds on 4/10 goals, showing that the closed-loop Imagine-then-Verify pipeline transfers to physical deployment, but remains vulnerable to appearance ambiguity and local drift. Across failed trials, a recurring failure mode emerges: the robot initially makes plausible progress, when faces alignment lost with key visual cues, it continues acting without recognizing the deviation.

Fig.~\ref{fig:real_world} shows a representative recovery case with SRM. Using navigation heuristics derived from simulation, the system detects the loss of alignment, identifies the last known goal-consistent region, and redirects the robot to recover. Although preliminary, this result suggests that the failure modes targeted by Robo-Cortex are not simulator-specific, and that these navigation heuristics can remain useful under real-world reflection.

%% file: 5_conclusion.tex
\section{Conclusion and Limitations}
\label{sec:conclusion}
We presented \textbf{Robo-Cortex}, a self-evolving embodied agent that reframes navigation from isolated action selection to continual strategy evolution. By combining imagine-then-verify planning, dual-grain cognitive memory, and autonomous knowledge induction, Robo-Cortex transforms embodied experience into reusable heuristics for future decision making. Across three embodied tasks, Robo-Cortex consistently improves success, efficiency, and generalization, while Robo-Cortex++ further shows that the learned strategy can continue to evolve through online memory and heuristic updating during inference. The transfer of induced heuristics to unseen environments, together with evidence from real-world robotic deployment, suggests that Robo-Cortex captures portable strategic structure rather than merely benchmark-specific behavior.

Despite these promising results, several limitations remain. Our current real-world validation is still preliminary and limited in scale, and broader physical experiments are needed to more fully assess robustness in diverse deployment settings. In addition, although the framework is intended to be model-agnostic, we have not yet systematically evaluated Robo-Cortex with different vision-language backbones or world models. Extending the evaluation across a wider range of VLMs and world models is therefore an important direction for future work.